\documentclass[conference]{IEEEtran}
\IEEEoverridecommandlockouts
\usepackage{cite}
\usepackage{amsmath,amssymb,amsfonts}
\usepackage{algorithmic}
\usepackage{graphicx}
\usepackage{textcomp}
\usepackage{xcolor}
\def\BibTeX{{\rm B\kern-.05em{\sc i\kern-.025em b}\kern-.08em
    T\kern-.1667em\lower.7ex\hbox{E}\kern-.125emX}}
\begin{document}

\title{Enhanced Audit Techniques Empowered by the Reinforcement Learning Pertaining to IFRS 16 Lease\\
}

\author{\IEEEauthorblockN{1\textsuperscript{st} Byungryul Choi}
\IEEEauthorblockA{\textit{Accounting Team, Hanjin Heavy Industries and Construction Co., LTD} \\
\textit{(Yonsei University)} \\
Seoul, South Korea \\
bryan.choi@yonsei.ac.kr}
}

\maketitle

\begin{abstract}
The purpose of accounting audit is to have clear understanding on the financial activities of a company, which can be enhanced by machine learning or reinforcement learning as numeric analysis better than manual analysis can be made. For the purpose of assessment on the relevance, completeness and accuracy of the information produced by entity pertaining to the newly implemented International Financial Reporting Standard 16 Lease (IFRS 16) is one of such candidates as its characteristic of requiring the understanding on the nature of contracts and its complete analysis from listing up without omission, which can be enhanced by the digitalization of contracts for the purpose of creating the lists, still leaving the need of auditing cash flows of companies for the possible omission due to the potential error at the stage of data collection, especially for entities with various short or middle term business sites and related leases, such as construction entities.

 The implementation of the reinforcement learning and its well-known code is to be made for the purpose of drawing the possibility and utilizability of interpreters from domain knowledge to numerical system, also can be called “gamification interpreter” or “numericalization interpreter” which can be referred or compared to the extrapolation with nondimensional numbers, such as Froude Number, in physics, which was a source of inspiration at this study. Studies on the interpreters can be able to empower the utilizability of artificial general intelligence in domain and commercial area. 
\end{abstract}


\section{Introduction}
Each implementation of International Financial Reporting Standards 16 Lease (“IFRS 16 Lease”) into individual companies was with various challenges due to the characteristic of the accounting standard requiring the analysis on the details of contracts for the purpose of distinguishing containment of (1) the existence of an identified asset, (2) the conveyance of the right to control the use of the identified asset, (3) the time period for the continuity of such right to control, (4) and the application of exemptions for the short-term lease and leases of low value assets. 

The IFRS 16 Lease incurs unprecedented changes in the financial statements, especially for the lease-heavy companies with the practice of asset-light, a well-known and prominent business operation strategy for the avoidance of risks in relation with the impairments and fluctuation on asset valuation and prices, such as shipping companies and construction companies that use many kinds of pricey machineries and services in the course of operating processes for the mega size projects, in return creating multiple layers of sub-contracts which can produce a large size of economic value, as well as a number of lease contracts that are subject to IFRS 16 newly implemented.

The digitalization of contracts and measurement of the right-of-asset and the liabilities according to cash outflow arrangements in future is one of the fair solutions for such companies with numerous lease contracts for the enjoyment of the advantage of computer engineering, calculation, and logical checkups that cannot be done if conducted manually, likewise, the adoption of artificial intelligence skills on such IFRS 16 lease accounting may shall be in need of study in consideration of its powerful convenience and proven performance in the matters of complexity. Such expectation is made among the industries as the artificial intelligence is under fast development, while direct and indirect of such techniques are not widely adopted with only small amount of reports of application, with many anticipations and comments that accounting will be surely under the course of such application due to the numeric characteristics of accounting practices, with possibility of indirect application of artificial intelligence, while direct application of artificial intelligence is still with some hindrances due to the need of more development for the logical accounting measurements.

An extrapolation is the process of evaluation of values in extended range or timeline through analysis of trend, widely adopted and accepted in contemporary accounting practice, especially for the purpose of fair market valuation. The extrapolation is not confined in a range of sole domain, however, is used for the purpose of the convergence of different domains such as between the geometry and architecture, or between the algebra and the economics. The equations based on nondimensionalized Froude Number is adopted for the purpose of calculation of the resistance of oceangoing vessels affected by numerous factors of friction of shell of the vessel and various responses of water movements and viscosity are one type of good examples of cross-domain extrapolation, since the developments base on such extrapolated equations erected numerous layers of industries on the ocean.

It is presumed that cross-domain extrapolation of the reinforcement learning is on the verge of becoming a necessity as the studies and performances on the reinforcement learning is on the course of development based on the computer games, while the applications and trials of the reinforcement learning on the industrial practices are assumed fairly limited. The reinforcement learning has been successfully able to reach out to the domain of games, sequential decisions, and series of spatial and chronological arrangements, all sum in the fairly tangible matters. 

The development of reinforcement learning is prone to focus on the matters that can have straightforward fair numeric reward values, not on the matter of evaluating blurred area which requires extensive understandings to make fair evaluation of the reward into numerical dimension, which is a stumbling block for the reinforcement learning can be extensively applied into some fields that require complex reasonings and comprehension. A fair calculation of the logics and reasonings into numerical dimension, however, can be a good start of the extensive application of the reinforcement learning as it has been so for centuries in the case of the nondimensional Froude Number effective to the real-world calculation of resistance of vessels.

\begin{equation}
Fr = \\\frac{u}{\sqrt{Lg}}
\end{equation}

\textit{u} : speed of flow

\textit{L}: characteristics length

\textit{g}: gravitation acceleration

 \bigskip

\section{Assessment on IFRS 16 Lease}
\bigskip

\subsection{IFRS 16 requiring analysis on contracts}

International Financial Reporting Standards 16 Lease (“IFRS 16 Lease”) adopted and released by the International Accounting Standards Board (IASB), effective on January first 2019 worldwide, has been considered a standard that requires developed intelligence technology skills due to its nature pertaining to the grounds of accounting which requires judgement on significant issues such as incentives, de facto lease periods, discount rates as per implicit rate of lessor or incremental borrowing rate of lessee, payments variations, detailed evaluation on non-lease elements, etc.

The contract digitalization might be one of the solutions for the rightful implement of IFRS 16 as the standard links directly to the characteristic of contracts, nevertheless, with some unjustifiable costs just for the implement of accounting aside the business practices unless the business prefers such digitalization, which is quite undesirable if the business has numerous short and middle term contracts and business fields such as a construction business holding a number of construction fields.\cite{b1}

In light of IFRS 16 requiring the projection of cashflows and the incremental borrowing rate (IBR) for the measurement of the lease liabilities and right-of-asset, and the actual cash outflows for the purpose of amortization of the lease liability, the characteristics of lease contracts showing various flexibilities on cash schedules and circumstances in the actual business environment shall require the cautious and watchful monitor not only on the details of the lease contracts but also on the behaviors of parties evolved and transactions occurred.

\subsection{Details of the completeness of IFRS 16}

It is still inevitable to have auditing on cash outflows for the completeness of lease contracts even for companies with the contract digitalization as some group of regular or repetitive cash outflows to specific counterparties may imply undetected or unrecorded contractual duties for the lease purpose with future cash outflows, which incurs tremendous time and costs if manually conducted.

With numericalization of cash outflows by the implementation of fair evaluation on the lease-likeness based cash outflows pattern and rightful translation into numerical values, as opted for creating language model and other deep learning area, the auditing on cash outflows for the purpose of lease completeness may be enhanced by the reinforcement learning, as the numericalized evaluation of lease-likeness can lead to election of sound rewards for the purpose of application of Bellman equation, which is also possible to be considered somewhat as the gamification of accounting, focusing on the characteristic of accounting, trying to archive the most rightful financial statements within the logical limitations and accounting standards.

Small medium enterprises without suitable ERPs may shall be in need of the reinforcement learning for the purpose of acquiring enough financial data out of accounting journals, while large enterprises with up-to-date ERPs with strong contract digitalization and cash integration suffer no difficulties on producing such financial data and subsequent accounting. 

\begin{figure}[htbp]
\centerline{\includegraphics[width=8cm]{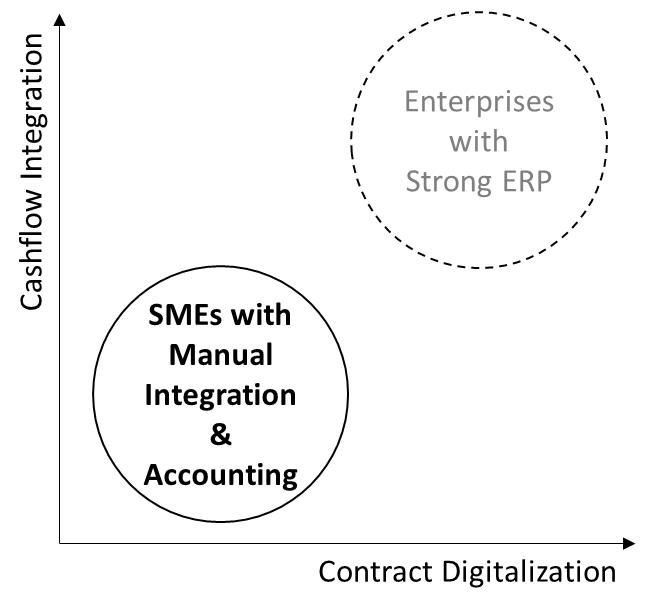}}
\caption{Entities in need of the reinforcement learning for lease completeness.}
\label{fig1}
\end{figure}

\bigskip

\textbf{Database: First Step for Information Produced by Entity}

If the collection of such lease contracts are made by purely email, the possibility of human error is unavoidable and the credibility of the information produced such manner is unreliable. Either the development of a new system collecting such information or providing manual entries to form a database system shall be considered. Either shall still be subject to the duty of providing the documents or proof of completeness for the purpose of collecting suitable data before identifying lease contracts. The completeness, as by its nature, shall be based upon seamless integration among the accounting data, or such proof if without such seamless database structured accounting system.

\bigskip

\textbf{Accounting Journals: Listings of Lease contracts
}

Repetitive cash outflows, not even the trails of cash outflows toward lessors and potential lessors, but also cash outflows with continuity to counterparties on a regular basis are subject to surveillance for the purpose of auditing completeness of lease contract listing since such regularity is a strong evidence of existence of lease contracts. The information produced by entity (IDE) pertaining to the list of lease contracts applied to IFRS 16 is to be classified into four categories, (1) IFRS 16 leases to be held to lease liabilities and right-of-use asset, (2) IFRS 16 leases to be described separately but not measured in the balance sheet due to its nature being short-term and low value lease, (3) leases no to be recognized but to be described for being finished within 12 months from the implementation of IFRS 16, and (4) leases that are not subject to the IFRS 16 definition. 

If the entities contains numerous unspecified contracts, in addition, potential errors on the entries shall be considered, as the small-medium sized real estate market such as housings are open to the individuals and the companies that are not tracking or containing sufficient information about such individuals without regulated policy or system shall conduct through analysis on entries based on similarities so that potential omission caused by bookkeeping error can be avoided. 

The review on cash outflows across the accounting codes incurs the need of review on all the cash out journals, amount payables, and related journals at least, applicable to numerous journals, which shall incur high expenses if manually conducted. The review on potential bookkeeping error across journals without sufficient information shall also require high expense if conducted, which is the reason why there is rare report of doing such audit on potential bookkeeping error. The reinforcement learning on each journal and case for such purpose shall be the best alternative that saves costs for audit and produces better information for IFRS 16 lease accounting.

\bigskip

\textbf{Amortization schedule and interest
}

As per IFRS 16, the discount rate for amortization schedule is outlined in this manner as follows: 

“At the commencement date, a lessee shall measure the lease liability at the present value of the lease payments that are not paid at that date. The lease payments shall be discounted using the interest rate implicit in the lease, if that rate can be readily determined. If that rate cannot be readily determined, the lessee shall use the lessee’s incremental borrowing rate.”

Lessee’s incremental borrowing rate, in addition, describes “The rate of interest that a lessee would have to pay to borrow over a similar term, and with a similar security, the funds necessary to obtain an asset of a similar value to the right of use asset in a similar economic environment.”

There were several opinions “similar term” reported among the accountants on the interpretation on the clauses above, mainly with three categories, at lease in South Korea back in 2019 at the stage of implement. (1) The interest rate shall be actual and proven interest rate that the company actually applied recently, which can be applied across assets and have possibility of ignoring the “similar security” and “similar economic environment” since each purchase of asset is subject to its own industrial environment. (2) The interest rate shall be calculated based on the weighted average period or duration period for the purpose of fulfill the “similar term”, and the interest rate shall be calculated in light of loss given default (LGD) as it was IFRS 9. (3) More than the weighted average period and LGD, the interest rate shall be measured in a complex model similar to financial bonds and be generated in a cautious and detailed manner. 

The more complexity on IBR calculation, the more the need of machine learning increased. SMEs with hundreds of lease contracts calculated on pure Excel or similar spreadsheet with interpretation of (2) and (3) above, shall be subject to the review on the accuracy of the calculation since its hyper complexity including, but not limited to, (1) cash schedule, (2) amortization table of liability, (3) amortization table of the right-of-use asset for deposit, restoration costs, initial direct costs, etc, (4) depreciation expense schedule, (5) weighted average period calculation, (6) basis of interest rate, such as national treasury coupon rate or bond market rate, (7) daily or monthly amortization, (8) existence of advance payments in various forms which is a widely accepted practice for lease contracts, (9) payment frequency and regularity, (10) unpaid or prepaid payments in real, as well as (11) foreign currency effects, in consideration of potential human error if manually conducted. 

\bigskip

\textbf{Consolidation and Understandings on Lease Contracts}

Considering the requirement of understandings on characteristics of lease contracts, and of generating various amortization schedules on each lease liability and right-of-use asset, as well as of adjusting the contracts applied to the consolidated statements within a group or such affiliation, the understanding of the company on the IFRS 16 lease standard also shall be subject to the audit of completeness, as this standard is more than numeric accounting, which is a right direction of accounting standard for the purpose of representing the financial statements close to the realities of the entity.

\subsection{Reinforcement learning for better auditing}

While one of the most detailed auditing practices back in 2019 at the implement stage was censoring all cash outflows in a same amount in to a same company at least twice, conducted solely by manual checkups, with tremendous cost and time. It may be praised for securing rock solid ground truth, but still lacks one thing, which is omitting the possibilities of bookkeeping mistakes. Auditing of cash outflow journals into one company at least twice without considering possibility of mistakes is comparably easier task than Auditing of cash outflow journals into one company at three times with one possible mistake, which produces multiple cases of mistake at least with three cashflows, wrong company names (with a number of candidates),  and wrong payments (with various reasons such as deduction, discount, netting, or comparably shorter residual time slot), surely something hard to be conducted manually but to be possible with well-designed machine learning tool such as reinforcement learning. While manually conducted, the cost of such auditing inclusive of such mistakes is unjustifiable and impracticable, if conducted properly with machine learning, the cost of such deeper auditing inclusive of possible mistake is much lesser than the manual auditing which leads to a firm justification of conducting such machine auditing.

The auditing inclusive of possible mistake is effective if the auditing period is focused on first quarter of fourth quarter of the financial period, as the periods is the most venerable to potential omission through simple mistake when such contracts start within the fourth quarter or ends within the first quarter, leaving the cash outflow journal difficult to detect some regular payments or potential contract duties, which leads to an implication that timely noise in accounting information within specific period can lead to faulted recognition of lease liabilities.

\bigskip

\section{Reinforcement Learning}

\bigskip

The reinforcement learning is a non-supervised machine learning seeking the maximum numerical reward on the basis of the Markov decision process (MDP) with the discrete probability distribution, or such circumstances, which requires the Agent, an important element of reinforcement learning with Action, Environment, State, and Reward, to learn the process to get to the maximum rewarding process through trial-and-error. 

Action is options of all possible moves or changes out of the state, and the state is a specific situation as per given Environment, which is a boundary and dimension for State, Action, Agent. The Reward is a given value for each for the pair of Action and State, which is a crucial element for the purpose of designing suitable reinforcement learning structure. 

Q-learning, the sum of two off-policy learning, the greedy policy and epsilon greedy, is often-used method as a developed reinforcement learning compared to the older version of reinforcement learning methods, with the proven effectiveness on seeking the best policy and trying other potential better options. The algorithm shall be described as below.\cite{b2}

\begin{equation}
Q(S, A) \gets  Q(S, A) + \alpha \cdot (R + \gamma \cdot \max_{a'} Q(S', a') - Q(S, A))
\end{equation}

The equitation (2) above means the memorization of rewards based on time diference, so that the possible sum of reward on each policy shall be made with discounted value as per the distance toward the future, so that iminant reward shall be preffred more than the same amount of the reward in the far future. The algorithm shows such step with elipson decay for the purpose of securing chance to make random trials within set range of possiblity.

Gridworld is well-known code as a representing case for the reinforcement learning since it reflects essential elements of the reinforcement learning. For the purpose of adopting reinforcement learning directly and of utilizing its efficient model code for auditing lease completeness, especially for the listing of existing lease contracts. 

Gridworld is a simple task of finding the shortest way, an useful example for reinforcement learning, whereby produces movements from a starting point to and endpoint on 4 × 4 or 5 × 5 grid, with only one notch movement allowed at once, to find the shortcut to get archive the goal, in this case reaching to the end point, evasive of hindrances if exists. The reinforcement learning adopts the Gridworld as an perfect model environment for Markov decision process (MDPs), with the key elements such as the policy that determines the probability of actions from each state, state-value function that discounted future expected rewards, and action-value function that returns expected reward of each action, and transition probability that moves state, and the reward function that specifies the reward of each action from state to state, which makes grounds for reinforcement learning controls.

 Q-learning, one of off-policy control widely opted for reinforcement learning due to its advantages of both extensive exploration of options and still learning for convergence, shall be elected as a control for such reinforcement auditing for the lease contract listing completeness inclusive of investigation of potential omission due to possible errors, deferrals to next year, or advance payment to the previous year.

\begin{figure*}[!t]
\centering{\includegraphics[width=14cm]{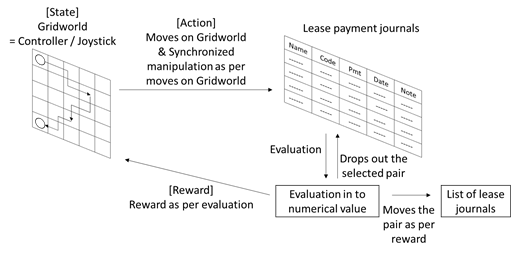}}
\caption{The reinforcement learning model for IFRS 16 lease.}
\label{fig2}
\end{figure*}

\subsection{Environment, state and action}

he connectivity between the movement on the Gridworld and modification of accounting journal, as if using the Gridworld as a joystick for changes on journals. If the modification has a suitable valuation system that returns right reward for the action and state, it shall be able to provide circumstances to practice the reinforcement learning. The creation of two pools, a nonnumeric value pool for corporate names and a numerical value pool for payments in this case, so that the options shall fit into the 2D Gridworld structure with each state shall be matched to each options, for example, Gridworld position (2, 3) means selection of second candidate from payments pool and third out of corporate names pool. Immediate evaluation of lease-likeness on journals produces the rewards, and possible lease candidates shall be transferred to lease listing, leaving the journals shortened, in the grid world viewpoint, shorter company name pools with shorter axis X. The remaining journals shall have the matching from the beginning, in this wise, the previous matching affects matching afterwards, producing diffusion decision model. The movement on the Gridworld follows the same rules in usual Gridworld, moving one step randomly among left, right, up, and down, which requires repetitive moves such as up-down-up-down or right-left-right-left to make same selection twice in the fastest way.

\begin{figure}[!ht]
\centering{\includegraphics[width=8cm]{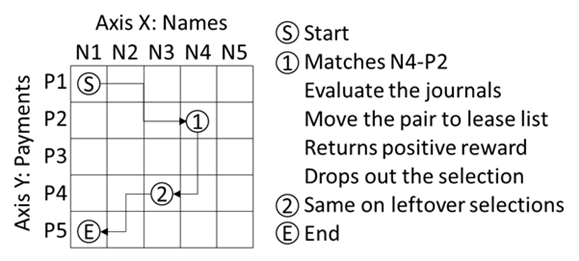}}
\caption{Example steps on the grid and consequent changes on journals.}
\label{fig3}
\end{figure}

\subsection{Regular reward and discount}

The purpose of the model is to list up the all the potential journals into the list, with specific steps, first, the classification of lease journals as they are, second, as a level 1 reinforcement learning implementation to shuffle nonnumeric item from each reasonable pool for each department or subdivision to find possible match for lease with a set of regular rewards which can also be referred as undiscounted rewards, third, as a level 2 reinforcement learning implementation to conduct the exchange of both nonnumeric item and numeric item, with discounted reward according to the coefficient of variation (CV) among the numeric values so that most even values with lowest CV can offer the greatest reward with minimized discount again the regular reward, for the purpose of minimization of the number of cases and prevent the curse of dimensionality with too many possible combinations.

\begin{figure}[!ht]
\centering{\includegraphics[width=8cm]{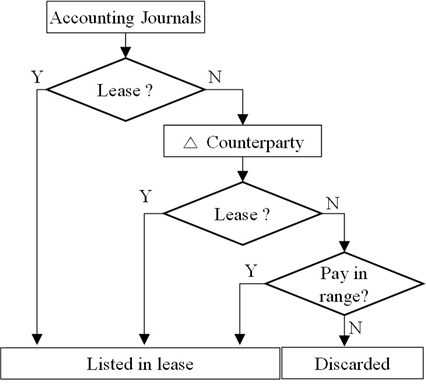}}
\caption{Assessment on accounting information by the model.}
\label{fig4}
\end{figure}

\subsection{Allowable coefficient of variance of lease payments}

The discounted reward according to the CV shall be estimated with the understanding of possible gap between the variance so that it may fit into the practice of accounting on the business. The possible maximum value of variance is as below, which implies that the unreasonable 50 times gap shall be possible unless a limitation on the allowance of CV is held. The possible allowance for gap among the payments shall be around 5 times, with near 0.7 CV, as per the prepayment practices and the possible payments in sums for several contracts with one company. The CV is the standard deviation of lease payments divided by average lease payments on same item in the same department at this study.

\begin{figure}[!ht]
\centering{\includegraphics[width=8cm]{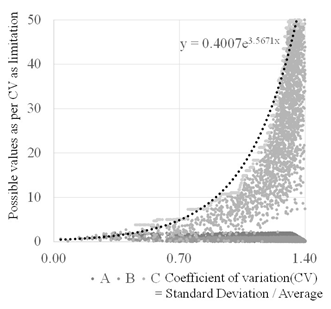}}
\caption{Allowable coefficient of variation.}
\label{fig5}
\end{figure}

\subsection{Negative reward and memory function for convergence}

The normal Gridworld sets disadvantage reward on each movements such as –1 or –0.1, however, -0.001 was set for the movement to make a neutral as possible reward for the purpose of reducing the any potential noise other than searching the most possible reward, but still with a bit of disadvantage so that it can learn to minimize the number of movements. Also articulate acceleration and bit of munipulation was made through building the call system, making a memory to remember a maxium reward to overcome next time, as  80,000+ iteration on each round even with the negative reward on movements without convergence unless such memory function is not made. This is the characteristic of the reinforcement learng that it is unable to learn if there are too much volatility and too less tendency on rewards. For the purpose of concentration on the construction of model as a whole, not on specific parts of the model, the effects of negetive reward is not considered at this study, even though the effect of negative reward is also an area that requires a detailed investigation.

\bigskip

\section{Result and Implications}

\bigskip

\subsection{Result with limitation of reinforcement learning}

The convergence for the highest reward needed accelerated epsilon decay for the purpose of aiding convergence, to reduce to time to archive the highest reward as soon as possible, similar to Convolution Neural Network (CNN) adopts early stopping for the models with overfitting. It required quite an endeavor to make right adjustment for the model to learn and reach to the convergence in the course of applying the reinforcement learning, which implies that the advanced learnings such as Deep-Q-Network (DQN) shall produce better results.

The convergence was made before trials of 50 times, model A with converged reward value with a bit of variance, and model B with stable converged reward value. For the purpose of auditing, model B is preferred as one stable reward means one best answer out of data.

\begin{figure}[!ht]
\centering{\includegraphics[width=8cm]{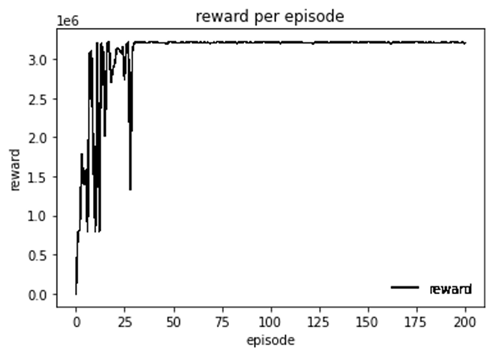}}
\caption{Plot of the convergence result of the model A.}
\label{fig6}
\end{figure}

\begin{figure}[!ht]
\centering{\includegraphics[width=8cm]{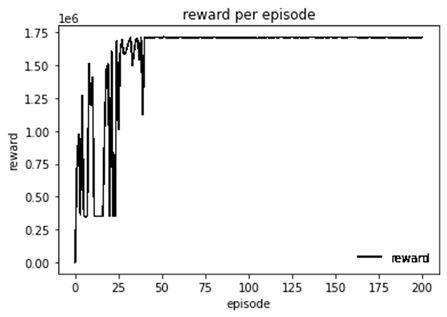}}
\caption{Plot of the convergence result of the model B.}
\label{fig7}
\end{figure}

The meaning of convergence means, at this study, that the maximum thorough review was made in light of potential human error at the stage of making enties about the lease cash out journals, which can increase the credibility of an unintegrated system without ERPs leaving accounting data to be under human serveilliance.

\begin{table*}[!t]
\caption{Company A: 1,300+ Contracts, 440,000+ journals, 80+ separate business sites}
\begin{center}
\begin{tabular}{||c c c c||}
\hline
                            & Manual + Email                          & Excel + Data Base                       & Reinforcement Learning      \\                                  \hline\hline
Task of Completeness check & Separately conducted by each department & Separately conducted by each department & Centered at Accounting Dept \\                                \hline
Lease Contract Listing     & Very Difficult                          & Half-manual                             & Automated                   \\  \hline
Potential Error Check      & Impossible                              & Very Difficult                          & Automated                   \\  \hline
IFRS 16 Identification     & Heavy Losses                            & Minor Losses                            & Near zero losses           \\  \hline
Cash Integration           & Impossible                              & Difficult                               & Easy                        \\                                 \hline
Working Hours              & 200 to 240 hours                        & 80 to 100 hours                         & 8 to 12 hours               \\                                  \hline
Double Check on Result     & Impossible                              & Difficult                               & Easy                        \\                                 
\hline
\end{tabular}

\end{center}
\end{table*}

\begin{table*}[!t]
\caption{Company B: 20+ Contracts with high stability}
\begin{center}
\begin{tabular}{||c c c c||}
\hline
                           & Manual + Email                          & Excel + Data Base                       & Reinforcement Learning \\ \hline\hline
Task of Completeness check & Separately conducted by each department & Separately conducted by each department & N/A                \\  \hline
Lease Contract Listing     & Stable, Easy                            & Stable, Easy                            & N/A                    \\  \hline
Potential Error Check      & Impossible                              & Impossible                              & N/A                    \\ \hline
IFRS 16 Identification     & Stable, Easy                            & Stable, Easy                            & N/A                    \\  \hline
Cash Integration           & Fine                                    & Fine                                    & N/A                    \\  \hline
Working Hours              & 15 to 20 hours                          & 15 to 20 hours                          & N/A                   \\  \hline
Double Check on Result     & Stable, Easy                            & Stable, Easy                            & N/A                    \\ 
\hline
\end{tabular}
\end{center}
\end{table*}

\subsection{Need of interpreter development and audit}

The reinforcement learning with interpretation intact with newly rectified with an accounting standard and its auditing, while the series of trials and studies are made with the expectation of building the artificial general intelligence, mainly based on the games of which actions, states, rewards can be described in numbers, implies the need of efforts from vice versa, i.e. domain dimensions to numeric dimensions, so that the artificial general intelligence may be applied to the domain even on the course of its own development. 

The development of series of interpreters that lets the artificial general intelligence or the reinforcement learning can be utilized in domain area shall be continued, and newly implemented accounting standard of IFRS 16 lease, can be one possible candidate domain for such purpose, as well as the audit practices on other accounting standards. As the purpose of audit is to understand the financial activities of companies, the reinforcement learning or other machine learning method can be utilized for better audit practices. Such machine based analysis can be used not only for the auditing purpose but also for financial advisory purpose, assisting the most reasonable decision making for the accountants and business management. Special groups of financial interpreters for artificial intelligence can also be developed into one sphere of business as the artificial intelligence itself develops.

The distinctive advantage of reinforcement learning was the enablement of the centralized completeness check task. Even the company B with 20+ lease contracts with high stability and small number of business sites were not able to conduct centralized completeness check based on the real cash outflows and shared the responsibility of completeness check on each department using lease contracts, with lingering question still whether that kind of completeness check shall be regarded “complete” since its credibility depends on the competence of departments in duty, not on the accounting information itself. 

On the other hand, the Company A with 1,500+ lease contracts with high volatility due to the nature of its business and 80+ business sites with continual installation and depreciation of a dozen of business sites annually were able to make centralized completeness check on 440,000+ accounting journals thanks to the implementation of the reinforcement learning, even with the additional check on potential errors that increases credibility of the lease accounting of the entity. Every facet of lease accounting, including but not limited to the lease contract listing, IFRS 16 lease identification, real cash outflow integration, and working efficiency was enhanced by the computing skills.

\bigskip

\section{Conclusions}

\bigskip

The implementation of both IFRS 16 lease and the reinforcement learning is the solution to reduce the repetitive tasks of checking and reviewing the lease transactions as well as to induce the possibility of system expansion of related functions since the formation of the lease accounting is not locked in the spread sheet but linked in the computer programming such as Python in the course of applying the reinforcement learning for the IFRS 16 lease. 

Some essentials of IFRS 16 Lease accounting at the stage of implementation are subject to human judgement especially for the classification as per own nature of lease contracts. The minimization of manual review shall be achieved to make the best judgement for the IFRS 16 lease standard, and for such purpose, the reinforcement learning is one of the best options as the study has revealed. 

Starting the completeness of IFRS 16 lease contracts, integrating the actual cash outflows, and redoubling the credibility through investigation on potential flaws, such implementation of the reinforcement learning shall be a good start for SMEs without expensive ERPs, as open source programming such as Python are easy to be acquired with near zero cost. The proof of completeness also can be easily produced in the course of building such programming. As the Company A shown above, a firm with accounting skills integrated with computing skills may produce better accounting information with low costs. 

Not only the reinforcement learning, but also many computing skills based on the artificial intelligence are on the stream. The accounting practices shall be also responsive to such changes so that industries can be mature and transparent enough to show the financial status as they really are, for the purpose of constructing sound capital market with right information.

\vspace{12pt}
\color{red}

\end{document}